\documentclass{llncs}

\usepackage{comment}
\usepackage{graphicx,wrapfig,lipsum}
\usepackage{multirow}
\usepackage{subcaption}
\usepackage{algpseudocode}
\usepackage{amssymb}
\usepackage{algorithm}
\usepackage{adjustbox}
\usepackage{float}
\usepackage{xcolor}

\usepackage{makeidx}  
\usepackage{etoolbox} 
\makeatletter
\patchcmd{\ps@headings}{\rlap{\thepage}}{}{}{}
\patchcmd{\ps@headings}{\llap{\thepage}}{}{}{}
\makeatother
\begin{document}
%
%

%
\mainmatter              
\title{LTS-NET: End-to-end Unsupervised Learning of Long-Term 3D Stable objects}
\titlerunning{Unsupervised Learning of 3D Stable objects}  
%
\author{Ibrahim Hroob \and Sergi Molina \and Riccardo Polvara \and \\
Grzegorz Cielniak \and Marc Hanheide }
\authorrunning{Ibrahim Hroob et al.} 
%
\tocauthor{Ibrahim Hroob, Sergi Molina, Riccardo Polvara, Grzegorz Cielniak,
Marc Hanheide}

\institute{
Lincoln Center for Autonomous Systems, University of Lincoln, UK\\
\email{ \{ihroob,smolinamellado,rpolvara,gcielniak,mhanheide\}@lincoln.ac.uk} }

\maketitle              

\begin{abstract}
    
    In this research, we present an end-to-end data-driven pipeline for determining the long-term stability status of objects within a given environment, specifically distinguishing between static and dynamic objects. Understanding object stability is key for mobile robots since long-term stable objects can be exploited as landmarks for long-term localisation. Our pipeline includes a labelling method that utilizes historical data from the environment to generate training data for a neural network. Rather than utilizing discrete labels, we propose the use of point-wise continuous label values, indicating the spatio-temporal stability of individual points, to train a point cloud regression network named LTS-NET. Our approach is evaluated on point cloud data from two parking lots in the NCLT dataset, and the results show that our proposed solution, outperforms direct training of a classification model for static vs dynamic object classification.

\end{abstract}

\section{Introduction}

    \begin{figure}[!ht]
        \centering
        \includegraphics[width=\textwidth, angle=0]{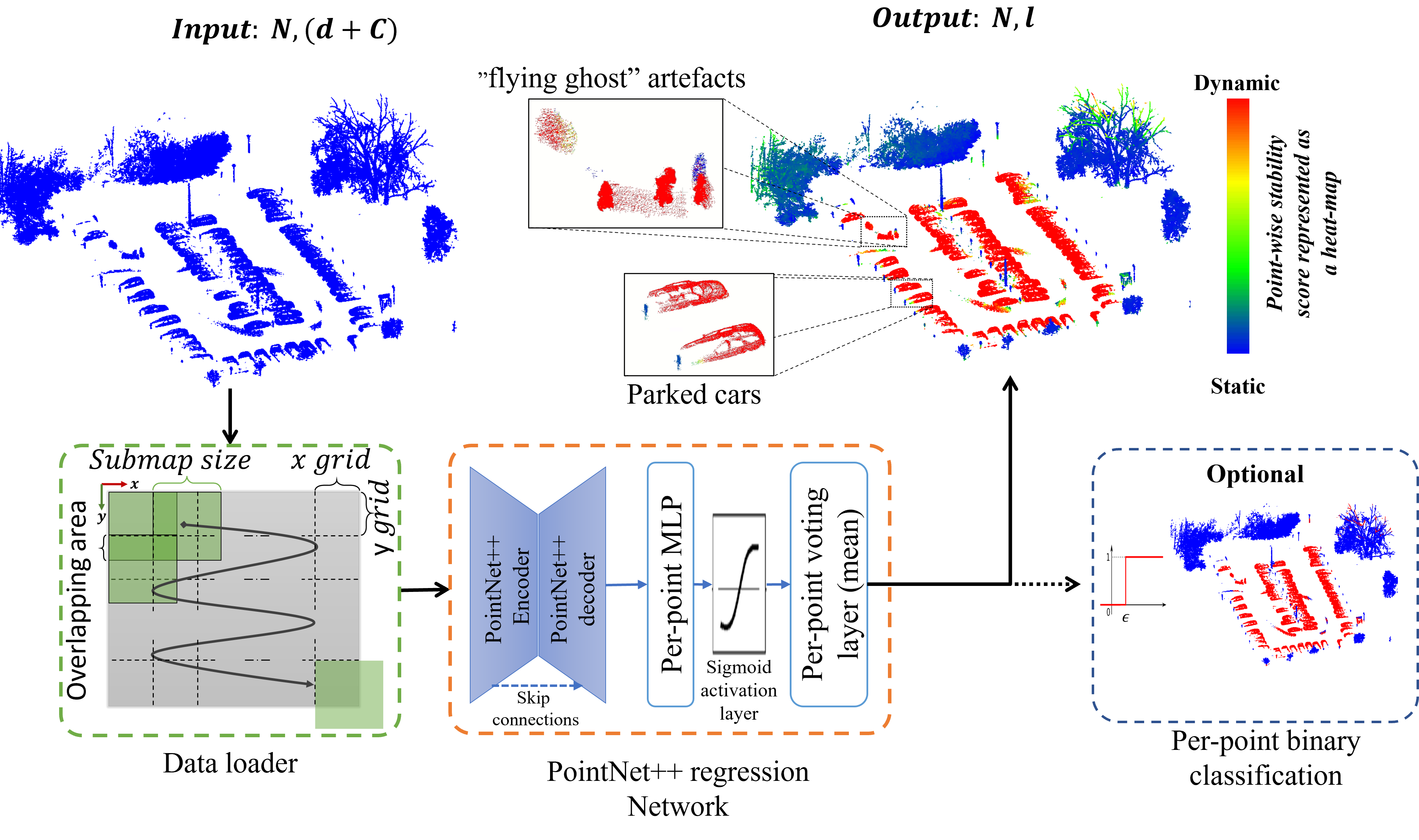}
        \caption{Overview of our proposed point-to-point regression model for estimating point-wise long-term stability in a 3D  point cloud map. 
        }
        \label{Fig:network}
    \end{figure}  

    The identification of object stability within a 3D point cloud of a given environment is crucial for mobile robots conducting long-term missions. These objects serve as landmarks for localization, particularly in environments that undergo significant changes over time, such as parking lots. Usually, the localization process is performed with respect to an internal representation of the environment, such as a 3D point cloud map for LiDAR-based localization, which represents a snapshot of the static state of the environment at the time of map acquisition \cite{pomerleau2014long}.
    
    However, the initial map may not be suitable for reliable long-term operations and can result in degradation of pose estimation for future sessions \cite{hong2022radarslam}. This is primarily due to two reasons: (i) the inclusion of dynamic objects as static, e.g. parked cars, even though such objects may appear static in the current scene, they are considered to be dynamic objects from the long-term perspective. (ii) The presence of "flying ghost" artifacts \cite{arora2021mapping} caused by moving objects while recording the data, such as pedestrians or cars in motion.

    Various solutions have been proposed for identifying the motion or stability status of objects, such as detecting and filtering dynamic objects using classical geometrical methods \cite{lim2021erasor,arora2021mapping,schauer2018peopleremover}. However, these methods primarily rely on motion information and may not detect dynamic objects that are stable in the current scene, such as parked cars. On the other hand, deep learning approaches can be used to identify object stability by achieving dense full-class segmentation \cite{dewan2017deep,zhou2018voxelnet,cortinhal2020salsanext}. These methods can be used to infer and detect object stability status, but they heavily rely on supervised annotated training data, which is not always available and can be costly to generate.

    In order to address the limitations of existing approaches, this paper proposes an end-to-end unsupervised learning method for determining the long-term stability status of objects within a 3D point cloud map. The output of this method is a point-wise spatio-temporal stability score, where higher values indicate that the point belongs to a dynamic object, such as a car or pedestrian, and lower values are associated to long-term stable objects such as buildings and trees. To achieve this, we propose an automatic labelling algorithm to generate training data with a point-wise labelling by utilizing different time instances of the environment. For the learning step, we propose a regression neural network based on the hierarchical PointNet++ \cite{qi2017pointnet} architecture.

    The main contributions of this research are three-fold: (i) an unsupervised automatic labelling algorithm that utilizes long-term observations of a given environment, (ii) LTS-NET a regression network based on PointNet++ for point-wise long-term stability score estimation, and (iii) a comprehensive evaluation of the proposed auto-labelling algorithm and regression neural network using real-world data, which demonstrates the effectiveness and convenience of the proposed approach as it does not require manual annotation.


\section{Related work}
\label{sec:related_work}

    Static and dynamic object segmentation is an active area of research, with methods broadly classified into geometry-based and deep learning approaches.

    Geometry-based methods are based on motion cues \cite{kim2020remove,pomerleau2014long}, ray tracing \cite{arora2021mapping,hornung2013octomap}, or voxel traversal \cite{schauer2018peopleremover}. Motion cues (visibility-based) approaches identify dynamic points by comparing the current laser scan with previous scans. For example, Pomerleau et al \cite{pomerleau2014long}. infer the dynamic part of a scene by comparing the incoming scan with a global map based on visibility assumptions. Ray tracing methods rely on shooting rays and checking for occlusions. These methods are typically computationally expensive and run offline. An example of a method using this strategy is OctoMap \cite{hornung2013octomap}, which is a probabilistic 3D mapping framework based on the octree data structure. The peopleremover \cite{schauer2018peopleremover} filters dynamic points using a voxel grid instead of an octree to store the identifier of all laser rays that hit the voxel.

    Deep learning approaches can be either supervised or unsupervised. Supervised method s\cite{zhou2018voxelnet,cortinhal2020salsanext,li2021multi} can achieve full classes semantic segmentation, but currently rely heavily on hand-annotated data and are prone to human error or unknown classes \cite{wong2020identifying}. Unsupervised methods are a more interesting choice for learning object semantics, usually in the form of dynamic or static binary objects. These methods are data-driven and require minimal or no supervision \cite{blum2022self}. For example, scene flow \cite{wang2021unsupervised,dewan2016rigid} approaches are being applied to point clouds directly in an unsupervised way to label points into moving or rigid objects between lidar frames. These methods are paired with a deep neural network for an end-to-end object semantic estimation.
    
    However, most methods mentioned above require motion information to infer dynamic objects; therefore, they cannot detect objects that can potentially move but are static in the current observation. In contrast to other works, our approach is more focused on identifying long-term stable objects in a given environment, as those objects are a key landmark to guarantee long-term localization without degradation in performance \cite{schaefer2019long}. Our method is unsupervised and does not require human input as it implicitly learns the long-term stable objects in an environment by exploiting previous temporal observations.

\section{Proposed method}

    In this paper, we propose a data-driven approach for identifying long-term stable objects in a given environment. We frame the problem as a regression task to associate a long-term stability score to each point in a point cloud. In particular, our model is able to predict that objects whose points have a high score are typically dynamic, we define a dynamic object as objects that can move on their own (e.g. cars, bikes, pedestrians, animals), while those with low-value score points are static, which are objects cannot move on their own (e.g. trees, poles, buildings).
    
    To accomplish this, we utilize a temporal sequence of 3D point cloud maps, denoted as $\mathbf{O}_{0:K}$, where $K$ is the total number of observations and $\mathbf{O}_{k}=\{ \mathbf{p}_{1},\dots,\mathbf{p}_{N}\}$ is the $k$-th observation ($k \in K$), with $N$ being the number of point clouds. Each point cloud $\mathbf{p}_{i} = \{x_i, y_i, z_i, Nx_i, Ny_i, Nz_i\}$ is defined by its 3D coordinates and associated surface normal vector.
    
    Our approach consists of two main steps: 1) designing an unsupervised labelling pipeline that assigns a stability score to each point cloud based on its spatial and temporal existence in all observations, where a score close to 0 indicates a long-term stable object and a score close to 1 indicates a dynamic object; 2) designing a regression network, denoted as $f(.)$, that can be trained on the stability scores to predict the stability of objects in a new given point cloud map.

    \subsection{Unsupervised point wise spatio-temporal labelling}
    \label{sec:Automatic_labelling}

    \begin{figure*}[!ht]
        \centering
        \includegraphics[width=\textwidth, angle=0]{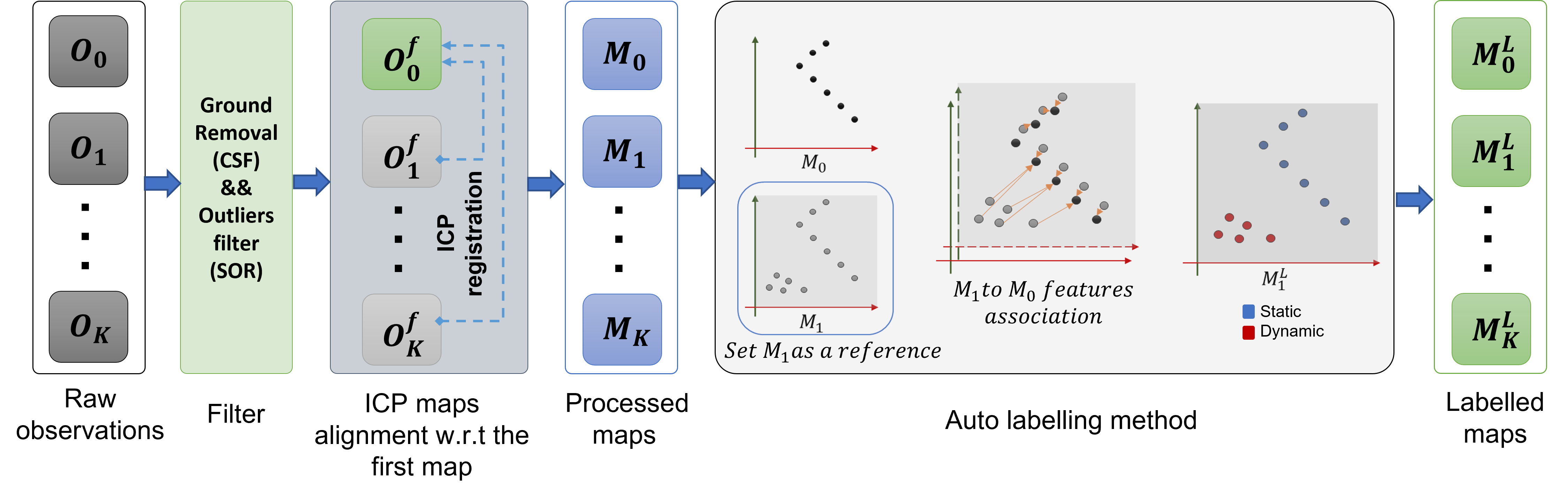}
        \caption{The unsupervised data labelling pipeline has two main parts: the data pre-processing block that filters the raw observations $\mathbf{O}_{0:K}$ and aligns them w.r.t the first map. Then the auto labelling method generates a point-wise stability score for each map by exploiting other maps.}
        \label{Fig:data_pipeline}
    \end{figure*}     

    In order to assign a stability score to each point in a cloud associated with an environment, we require at least two observation of the same environment. The process of generating the stability score is illustrated in Figure~\ref{Fig:data_pipeline}. The first step of the labelling pipeline is to filter the observations. This is achieved by using the Cloth Simulation Filtering (CSF) algorithm \cite{arora2021mapping} to remove the ground plane points, resulting in the set of off-ground points $\mathbf{O}_{k}^{offground}$. The motivation for removing the ground plane points is threefold: (i) the ground plane is stable, (ii) it increases the disparity of point labels when extracting the spatial distance to the nearest neighbour point in other observations/maps at later stages, and (iii) it reduces the overall map size. After applying CSF, we remove the outliers using the Statistical Outlier Removal (SOR) filter \cite{rusu2007towards} from the off-ground points. The output of this step is $\{\mathbf{O}_{0}^{f}, \dots, \mathbf{O}_{K}^{f}\}$ indicating the filtered observations. 
    
    The next step is to register the filtered observations with respect to the first observation. This step is crucial for accurately associating features between all filtered observations. To perform the registration, we use the Iterative Closest Point (ICP) algorithm to find the best transformation matrix $T_{\mathbf{O}_{k}^{f}}^{\mathbf{O}_{0}^{f}}$, which is then used to transform the observation as follows:

    \begin{equation}
    \mathbf{M}_{k} =  T_{\mathbf{O}_{k}^{f}}^{\mathbf{O}_{0}^{f}} \cdot \mathbf{O}_{k}^{f}
    \end{equation}    
    
    where $\mathbf{M}_{k}$ is the transformed observation with respect to the first observation, and we note that $\mathbf{M}_{0} = \mathbf{O}_{0}^{f}$.
    
    After pre-processing the data, the spatio-temporal label for each point $\mathbf{p}_{i} \in \mathbf{M}_{k}$ is determined as follows: we first find a vector $\vec{d}$ of the spatial distance to the closest point in all other observations $\{\mathbf{M}_{j}\}_{j=0}^{K}, j \neq k$, where for example, $d_{0}$ is the spatial distance to the closest point $\mathbf{q}$ in $\mathbf{M}_{0}$, and the size of $\vec{d}$ is $K$. Then, a point label is assigned as:
    
    \begin{equation}
    l_{i} = 1 - e^{-\lambda . \max(\vec{d})}
    \end{equation}
    
    where $\lambda$ is a hyperparameter that controls the labelling sensitivity. We use the maximum spatial distance across the temporal slices as a stability feature because a dynamic object may not appear or change location in one of the temporal slices resulting in a larger spatial distance feature associated with it. While on the other hand, stable objects will appear in all slices in the same location, resulting in a small distance feature. Additionally, we use the Cumulative Distribution Function (CDF) of an exponential function to map the spatial distance into CDF space and bound the value between 0 and 1.

    \subsection{Regression Network} 
    The stability labels generated in this work are continuous values that range between 0 and 1, which is not compatible with classical segmentation or binary classification networks that expect discrete labels. While it is possible to add a threshold to the input data to split the labels into two states, we are interested in directly learning and making use of the stability labels. To accomplish this, we propose a regression network, called LTS-NET, that can learn the stability labels directly from point cloud data.
    
    LTS-NET is based on the pioneering work of PointNet++ \cite{qi2017pointnet}, which is composed of an encoder-decoder structure with multiple layers to enable the learning of spatial features at different scales. The encoder layers are referred to as abstraction layers (AL) and the decoder layers are referred to as feature propagation layers (PL) as introduced in the original implementation of PointNet++. For LTS-NET, we use 5 AL and 5 PL. The input number of points for each AL is as follows: $N_1: 1024$, $N_2: 512$, $N_3: 256$, $N_4: 128$ and $N_5: 32$ with the following sampling radius at each layer $r_1: 0.1$ m, $r_2: 0.2$ m, $r_3: 0.4$ m, $r_4: 0.8$ m and $r_5: 1.4$ m. In the output layer, we used the Sigmoid activation function to bound the estimates between $[0,1]$; It is worth stating that we have also experimented with other activation functions such as CDF of an exponential function, but found that they gave similar results. This is likely due to the weights of the network adapting to the output activation layer.
    
    \textbf{Regression Labels imbalance: }
    To address the imbalance in continuous labels, we adopted a sample weighting approach solution proposed by Steininger et al. \cite{steininger2021density}. The weight for each sample is based on the rarity of the label, so the weight is inversely proportional to the probability of the label occurrence. This will help the model to better estimate the rare cases \cite{krawczyk2016learning}. The weighting function is defined as
    \begin{equation}
    \label{eq:dense_weight}
        f_w(\alpha,y) = \frac{\max(1 - \alpha p'(y), \epsilon)}{ \frac{1}{N} \sum_{i=1}^{N} (\max(1 - \alpha p'(y_i), \epsilon)) },
    \end{equation}
    where $p$ is the target variable density function, $N$ is the number of data points, $y = {y_1, ... y_N}$ is the target values, $p' = (\ p(y)-\min(p(y))\ ) / (\ \max(p(y))-\min(p(y))\ )$ is the normalized density function \( \in [0,1] \) , hyperparameter \( \alpha \in [0, \infty) \) which emphasize the weighting scheme, and \( \epsilon \) is a small positive real number to avoid negative or 0 weights. For more details and experiments of the effectiveness of this weighting scheme, we refer to the original density‑based weighting article~\cite{steininger2021density}.

    \textbf{Regression Loss function: }
    We apply the weighted Root Mean Square Error to supervise the spatio-temporal score prediction for LTS-NET:
    \begin{equation}
    \label{eq:RMSE}
        \mathcal{L} = (\frac{1}{N} \sum_{i=1}^{N} f_w(\alpha,y_i)(\hat{y_i} - y_{i})^2)^{\frac{1}{2}}.
    \end{equation} 
    Combining the sample gradients with the weight $f_w(\alpha,y_i)$ leads to larger gradients for the rare cases; in this way, the model is forced to have better estimates for the rare values as discussed in \cite{steininger2021density}. 
    
    \textbf{Data loader: }
    When receiving in input a map, the first layer of LTS-NET subsample a fixed number of points in order to make the learning task more tractable. Because of this, it makes inconvenient to feed the entire map at once, because the local geometry of the environment will be lost. In order to keep the network's input representative enough, we divide the original map into multiple submaps and we feed them to the network in an iterative fashion. Choosing the appropriate submap size and number of points is a challenging problem due to the uneven distribution of data and the scale of features. In this work, we used a submap size of $10\times10$ m in the x and y axis, with no constraints in the z axis, and a number of points set to $4096$, we found those values empirically while designing and testing the model. The submaps are generated in a convolutional way by moving with a fixed grid in the x and y axis (as illustrated in Fig.~\ref{Fig:network} data-loader block). The approach guarantees full map coverage and captures all features. The grid size for our experiments is $50\%$ of the submap size, which gives $50\%$ overlap.
    
    \textbf{Voting layer: }
    At the inference stage of the network, some points may be assigned multiple predictions due to overlapping submaps. In PointNet \cite{qi2017pointnet}, they use a voting pool, where the class that gets more votes will be the point label. However, in regression, the prediction value is continuous; therefore, we take the mean of predictions to get the point estimated stability score.   

\section{Experiments}

    \subsection{Dataset}
        \label{sec:dataset}
        In this study, we evaluate our proposed approach using the North Campus Long-Term (NCLT) dataset \cite{carlevaris2016university}. The NCLT dataset is well-suited for our purposes as it was collected over a 15-month period, consisting of 27 recordings, providing a rich source of data for long-term operation analysis. The data was collected using a two-wheeled robot equipped with a Velodyne HDL-32E LiDAR, wheel encoders, GPS, IMU, and a gyroscope, on the University of Michigan campus.

\begin{figure}[!ht]
        \centering
            \includegraphics[width=0.8\textwidth]{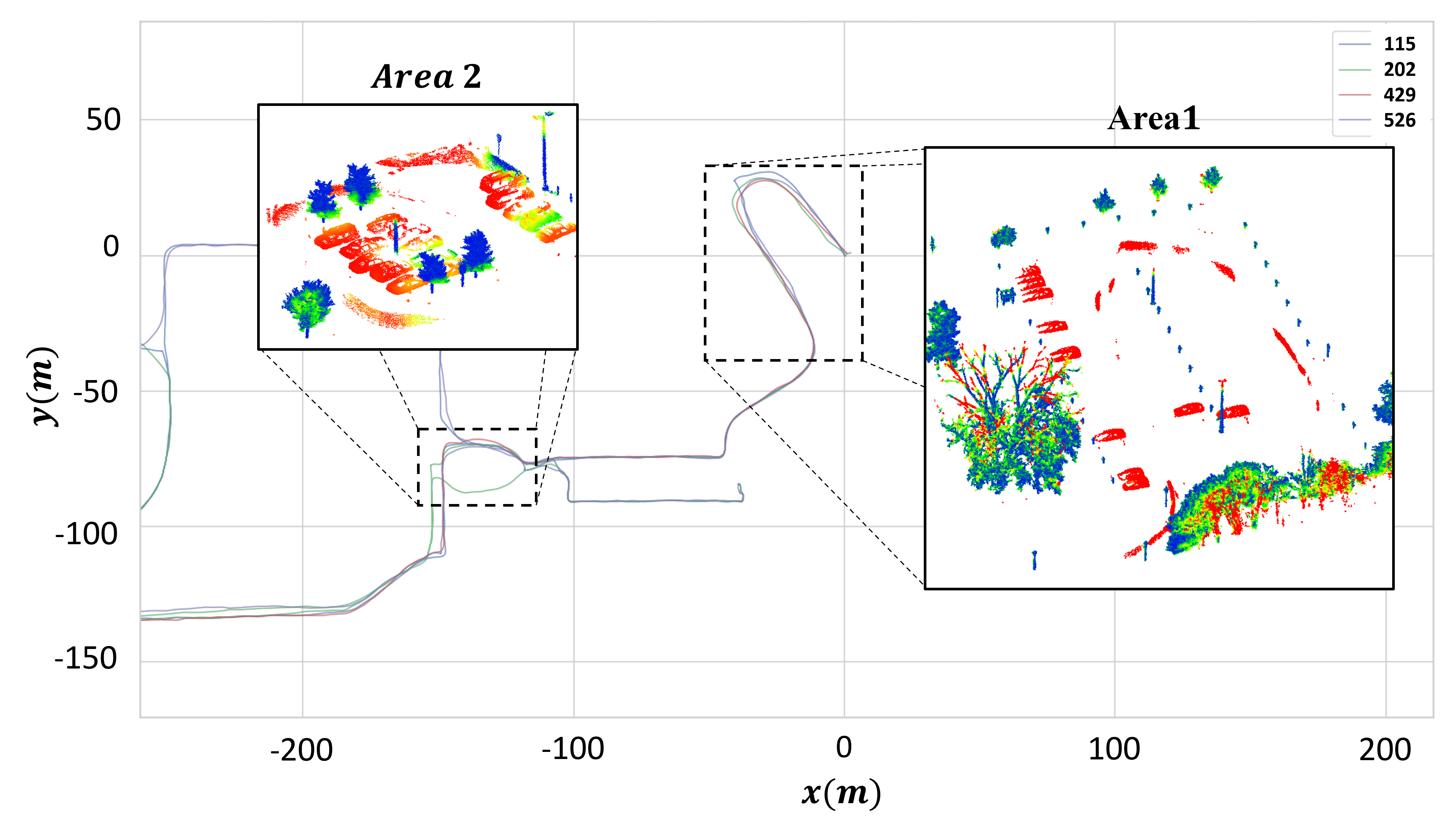}
            \caption{ The two areas in NCLT dataset that we used to train and evaluate our approach   }\label{fig:nclt_maps}
        \end{figure}

        To construct the 3D point cloud maps for our experiments, we employed the Simultaneous Localization and Mapping (SLAM) system FAST-LIO \cite{xu2022fast}, which only requires 3D LiDAR data from the Velodyne and IMU data. From the 27 available data recording sessions, we selected 5 recordings that had the most overlap, specifically: 2012-01-15, 2012-02-02, 2012-04-29, 2012-05-26, and 2012-08-04. Within these observations, we focused on two areas, as shown in Fig.~\ref{fig:nclt_maps}, which are two parking lots. The size of Area 1 is 70$\times$90 m, and Area 2 is 40$\times$30 m. The raw observations are denoted as $\{\mathbf{O}_{0}, \dots, \mathbf{O}_{4} \}$, and the processed observations as ${\{\mathbf{M}_{0}, \dots, \mathbf{M}_{4}\}}$ throughout the rest of the manuscript.

        \textbf{Ground Truth Binary Maps for Object Stability: }
        to evaluate the quality of the auto-labelled maps and the predictions of LTS-NET, it is necessary to use a ground truth binary labelled map as a reference. However, long-term object stability annotations are not available for the NCLT dataset. Therefore, we manually generated ground truth maps using the CloudCompare software\footnote{CloudCompare (version 2.12) [GPL software]. (2022). Retrieved from http://www.cloudcompare.org/}. The classification was performed as follows: trees, light posts, and poles were labelled as stable objects, while all other objects were labelled as dynamic.

        \textbf{Generating stability labels:}
        in order to generate the stability labels for each map in the experimental area, we selected a reference map and then generated the stability labels with respect to all other maps in the experimental area by using the method described in Section \ref{sec:Automatic_labelling}. For example, the stability labels for $\mathbf{M}_{1}$, is found w.r.t $\{\mathbf{M}_{j}\}_{j=0}^{K}, j \neq 1$. It is worth noting that the temporal sequence of the maps is not important for identifying the static objects as they are not affected by the order in which the observations are being processed. 

    \subsection{Baseline}
        To demonstrate that the proposed approach of using a regression network to learn a stability score for point cloud data outperforms traditional binary classification methods, we used the PointNet++ network as a baseline. The baseline is a PyTorch implementation of PointNet++ \footnote{https://github.com/yanx27/Pointnet\_Pointnet2\_pytorch} that uses the weighted negative log-likelihood loss, where the weights are based on the distribution of the classes. Additionally, we used the same submap size to generate data for both the baseline and the proposed regression model. The baseline was trained and evaluated on the ground truth binary data, while the regression model was trained on the stability labels generated by the unsupervised auto-labelling method.

    \subsection{Experimental scenarios and setup}
        \label{sec:eval_network}
        We conducted two experiments: Experiment $1$ and Experiment $2$. For Exp.$1$, both models were trained on the labelled map $\mathbf{M}_0^L$ of Area $1$ and evaluated in all other maps. The second Exp is similar to the first one, but both models were trained on $\mathbf{M}_0^L$ of Area $2$. The motivation behind these experiments is twofold: first, we want to assess the spatio-temporal generalization capability of the network, to evaluate if LTS-NET can infer stable object at different time slice of the same environment; second, we want to assess the generalization capability in unseen environments, which were not part of the training set.
        
        To compare the output of LTS-NET with a classification mode, we converted the regression output into binary classes using a threshold value $\epsilon$. An optimal threshold for the regression model can be found with the Receiver Operating Characteristic (ROC) curve, which could be found by minimizing or maximizing a certain metric \cite{zou2013optimal}. In our case, the metric that we are trying to maximize is the geometric mean \cite{lawson2001geometric}, which is a metric for imbalanced classification that, if optimized, gives a balance between the sensitivity (true positives rate) of the model and the specificity (inverse of false positive). 

        The optimal threshold $\epsilon$ for Exp.$1$ is found using the ROC curve of the inferred labels of $\mathbf{M}_0$ of Area $1$, which is equal to $0.269$ corresponding to $0.626$ m . Then, for consistency and to not over-fit the results, we used this value to convert the regression output to binary for all evaluated maps in Exp.$1$. Those binary labels are evaluated w.r.t the ground truth labels. For Exp.$2$, the optimal threshold was found for ($\mathbf{M}_0$) of Area $2$ that is equal to $0.3593$ ($0.89$ m), then we did the same as for Exp.$1$.

        \textbf{Implementation: } We train and evaluate both the binary classification (baseline) and LTS-NET on a workstation with Intel Core i7-6850K CPU, 64GB RAM and an NVidia GTX 1080ti GPU with 12 GB RAM. The model is implemented using PyTorch framework. We used a learning rate of $0.001$, momentum $0.9$, and trained for $60$ epochs for both the baseline and our model (regression). The training time for both models was 4 hours for training on $\mathbf{M}_{0}$ of Area $1$ and around 2 hours for training on $\mathbf{M}_{0}$ of Area $2$.

        \textbf{Metric: }
        The metric used to evaluate the baseline segmentation model and the thresholded values of the regression model is the mean intersection over union (mIoU), defined as $mIoU = \frac{1}{N} \sum_{c=0}^{N}IoU_c$, where $N$ is the number of classes, $IoU_c = ( |\mathbf{p}_c \cap \mathbf{G}_c|)/(\mathbf{p}_c \cup \mathbf{G}_c)$, $c$ is the point class (stable/unstable), $\mathbf{p}_c$ is the predicted set, $\mathbf{G}_c$ is the ground truth set.

        To evaluate LTS-NET predictions w.r.t to the auto labelled data, we used Root Mean Square Error (RMSE) metric expressed as $RMSE =  (\frac{1}{N} \sum_{i=1}^{N} (\hat{l_i} - l_{i})^2)^{\frac{1}{2}}$, where $N$ is the number of labels (points) and $\hat{l_i}$ is the predicted label. 
        
    \subsection{Results on NCLT parking lot areas}
        \subsubsection{Evaluating the unsupervised labelling}  \label{sec:eval_unsupervised_labelling}
        To evaluate the accuracy of the auto-labelled data, we use the area under the ROC curve, known as ROC AUC, which summarizes the performance of the auto-labelling by a single number with values between $0.5$ (random labelling) and $1.0$ (perfect labelling). The ROC curves are computed by comparing the stability scores with the ground truth binary data at different thresholds. Tab.~\ref{table:auto_label_auc} summarize ROC AUC for Areas $1$ and $2$, which indicates a good performance of the auto-labelling algorithm. 
        \begin{table}[ht]
        \caption{Auto labeling data noise evaluation using the area under ROC curve}
        \label{table:auto_label_auc}
        \centering
        \begin{tabular}{|c|ccccc|}
        \hline
        Map & $\mathbf{M}_0$ & $\mathbf{M}_1$ & $\mathbf{M}_2$ & $\mathbf{M}_3$ & $\mathbf{M}_4$ \\ \hline
        ROC AUC Area 1 & 0.997 & 0.999 & 0.995 & 0.997 & 0.988 \\ \hline
        ROC AUC Area 2 & 0.999 & 0.999 & 0.999 & 0.999 & 0.999 \\ \hline
        \end{tabular}
        \end{table}

        \subsubsection{Evaluating maps inference}
\begin{figure}[!ht]
        \centering
            \includegraphics[width=\textwidth]{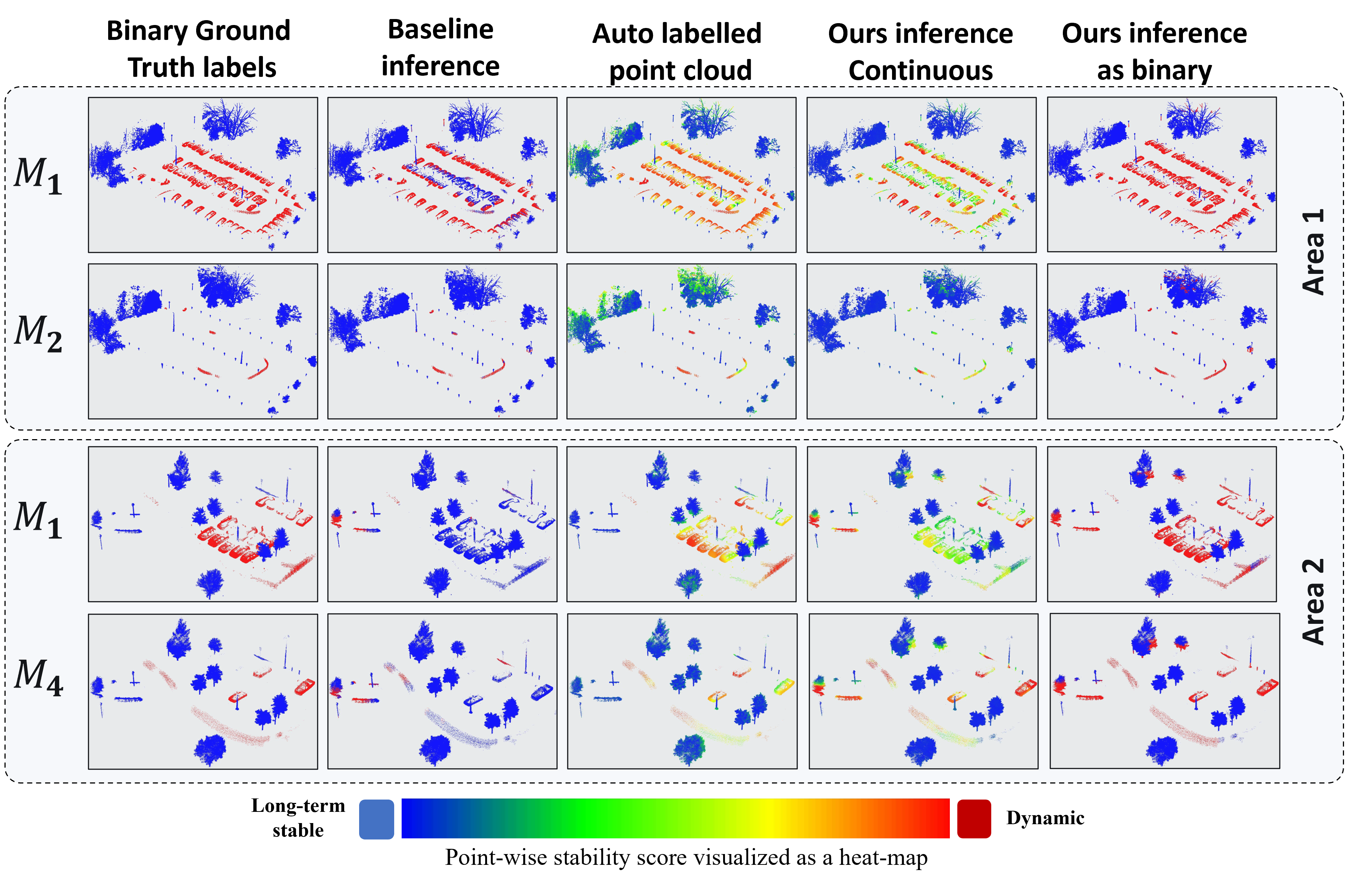}
            \caption{ Visual evaluation of the inferred maps of both areas in Exp.1, this visualization is only for the best and worst results for the Exp.}\label{fig:tests}
        \end{figure}

        As shown in Tab.~\ref{table:basline_vs_ours}, when both models were trained and evaluated in the same area, they showed a comparable performance despite the fact that LTS-NET was trained on the unsupervised labels only. The evaluation for both models is w.r.t the ground truth labels. 
        The regression output was converted to binary using the optimal threshold for each Test as explained in Sec.~\ref{sec:eval_network}. 
        However, the interesting results are when evaluating the models in the opposite area used for training. For instance, in Exp.$1$, LTS-NET outperforms the binary classification model by a large margin in most maps; on average, the mIoU score improved by $34.2\%$ over the baseline. For Exp.2 on Area $1$, the regression model shows improvement over the baseline with an average increase by $14.6\%$ in the mIoU score. However, when training on Area 2 and testing on Area 1 we obtain a significant lower score (still superior to the one of the baseline). This is due tot he fact that Area 2 is smaller than Area 1, presenting way fewer feature from which the network can learn from. Overall, the results confirm that the continuous labels can better utilize the 3D spatial information in the point cloud data. A visualization of the best and worst results of Exp.$1$ are shown in Fig.~\ref{fig:tests}.

        \begin{table*}[ht]
        \centering
        \caption{Performance comparison between the baseline and our approach, the metric used to compare both models is mIoU expressed in \%. The baseline is trained and evaluated on the ground truth data. LTS-NET is trained in an unsupervised fashion and is evaluated w.r.t the ground truth binary labels. In addition, the evaluation regression RMSE loss is presented. The '---' indicates training data.}
        \label{table:basline_vs_ours}
        \begin{tabular}{ccc|ccccc|ccccc|}
        \cline{4-13}
        \multicolumn{1}{l}{} & \multicolumn{1}{l}{} & \multicolumn{1}{l|}{} & \multicolumn{5}{c|}{\textbf{Area 1}} & \multicolumn{5}{c|}{\textbf{Area 2}} \\ \hline
        \multicolumn{1}{|c|}{Exp} & \multicolumn{1}{c|}{Model} & Metric & $\mathbf{M}_0$ & $\mathbf{M}_1$ & $\mathbf{M}_2$ & $\mathbf{M}_3$ & $\mathbf{M}_4$ & $\mathbf{M}_0$ & $\mathbf{M}_1$ & $\mathbf{M}_2$ & $\mathbf{M}_3$ & $\mathbf{M}_4$ \\ \hline
        \multicolumn{1}{|c|}{\multirow{3}{*}{Exp.1}} & \multicolumn{1}{c|}{Baseline} & mIoU & --- & 0.85 & \textbf{0.85} & \textbf{0.89} & \textbf{0.92} & 0.36 & 0.33 & 0.34 & 0.48 & 0.47 \\ \cline{2-3}
        \multicolumn{1}{|c|}{} & \multicolumn{1}{c|}{\multirow{2}{*}{LTS-NET [ours]}} & mIoU & --- & \textbf{0.98} & 0.73 & 0.88 & 0.86 & \textbf{0.78} & \textbf{0.82} & \textbf{0.78} & \textbf{0.66} & \textbf{0.65} \\ \cline{3-3}
        \multicolumn{1}{|c|}{} & \multicolumn{1}{c|}{} & RMSE & --- & 0.15 & 0.10 & 0.12 & 0.13 & 0.26 & 0.25 & 0.25 & 0.20 & 0.22 \\ \hline
        \multicolumn{1}{|c|}{\multirow{3}{*}{Exp.2}} & \multicolumn{1}{c|}{Baseline} & \multicolumn{1}{c|}{mIoU} & 0.44 & 0.28 & 0.50 & 0.49 & 0.49 & --- & \textbf{0.95} & 0.96 & 0.84 & \textbf{0.86} \\ \cline{2-3}
        \multicolumn{1}{|c|}{} & \multicolumn{1}{c|}{\multirow{2}{*}{LTS-NET [ours]}} & mIoU & \textbf{0.66} & \textbf{0.54} & \textbf{0.55} & \textbf{0.57} & \textbf{0.61} & --- & \textbf{0.95} & \textbf{0.97} & \textbf{0.86} & \textbf{0.86} \\ \cline{3-3}
        \multicolumn{1}{|c|}{} & \multicolumn{1}{c|}{} & RMSE & 0.21 & 0.37 & 0.15 & 0.15 & 0.15 & --- & 0.16 & 0.17 & 0.11 & 0.12 \\ \hline
        \end{tabular}
        \end{table*}
        
\section{Conclusion}
\label{sec:conclusion}
    In this paper, we have proposed a novel end-to-end unsupervised deep learning method for estimating the long-term stability of objects in a 3D point cloud map. Our approach consists of two parts: an unsupervised labelling algorithm that generates a point-wise stability score by utilizing the temporal observations of a given environment, and LTS-NET, a stability point-wise regression network based on PointNet++ that is trained on the stability labels and can be used to infer objects' stability in similar environments with no previous observations.
    
    Experimental results have shown that the proposed method can efficiently identify which points in a map belong to long-term stable objects (i.e. poles, tree trunks, buildings), which can improve long-term localization in environments that are subject to continuous changes. Additionally, our method has revealed that long-term stable object classification is best performed by training a regression model on stability scores, followed by thresholding, compared to directly training a binary classifier. To the best of the authors' knowledge, this has not been previously investigated and reported in the literature. 
    
    As future work, we plan to explore extracting long-term stable objects directly from individual 3D LiDAR scans. We also aim at using stable feature maps obtained by LTS-NET for evaluating and improving upon long-term robot localization.

\section*{Acknowledgements}
    This work has been supported by the European Commission as part of H2020 under grant number 871704 (BACCHUS).

\bibliographystyle{splncs}
\bibliography{ref}

\begin{thebibliography}{10}

\bibitem{pomerleau2014long}
Pomerleau, F., Kr{\"u}si, P., Colas, F., Furgale, P., Siegwart, R.:
\newblock Long-term 3d map maintenance in dynamic environments.
\newblock In: 2014 IEEE International Conference on Robotics and Automation
  (ICRA), IEEE (2014)  3712--3719

\bibitem{hong2022radarslam}
Hong, Z., Petillot, Y., Wallace, A., Wang, S.:
\newblock Radarslam: A robust simultaneous localization and mapping system for
  all weather conditions.
\newblock The International Journal of Robotics Research (2022)
  02783649221080483

\bibitem{arora2021mapping}
Arora, M., Wiesmann, L., Chen, X., Stachniss, C.:
\newblock Mapping the static parts of dynamic scenes from 3d lidar point clouds
  exploiting ground segmentation.
\newblock In: 2021 European Conference on Mobile Robots (ECMR), IEEE (2021)
  1--6

\bibitem{lim2021erasor}
Lim, H., Hwang, S., Myung, H.:
\newblock Erasor: Egocentric ratio of pseudo occupancy-based dynamic object
  removal for static 3d point cloud map building.
\newblock IEEE Robotics and Automation Letters \textbf{6}(2) (2021)  2272--2279

\bibitem{schauer2018peopleremover}
Schauer, J., N{\"u}chter, A.:
\newblock The peopleremover—removing dynamic objects from 3-d point cloud
  data by traversing a voxel occupancy grid.
\newblock IEEE robotics and automation letters \textbf{3}(3) (2018)  1679--1686

\bibitem{dewan2017deep}
Dewan, A., Oliveira, G.L., Burgard, W.:
\newblock Deep semantic classification for 3d lidar data.
\newblock In: 2017 IEEE/RSJ International Conference on Intelligent Robots and
  Systems (IROS), IEEE (2017)  3544--3549

\bibitem{zhou2018voxelnet}
Zhou, Y., Tuzel, O.:
\newblock Voxelnet: End-to-end learning for point cloud based 3d object
  detection.
\newblock In: Proceedings of the IEEE conference on computer vision and pattern
  recognition. (2018)  4490--4499

\bibitem{cortinhal2020salsanext}
Cortinhal, T., Tzelepis, G., Erdal~Aksoy, E.:
\newblock Salsanext: Fast, uncertainty-aware semantic segmentation of lidar
  point clouds.
\newblock In: International Symposium on Visual Computing, Springer (2020)
  207--222

\bibitem{qi2017pointnet}
Qi, C.R., Su, H., Mo, K., Guibas, L.J.:
\newblock Pointnet: Deep learning on point sets for 3d classification and
  segmentation.
\newblock In: Proceedings of the IEEE conference on computer vision and pattern
  recognition. (2017)  652--660

\bibitem{kim2020remove}
Kim, G., Kim, A.:
\newblock Remove, then revert: Static point cloud map construction using
  multiresolution range images.
\newblock In: 2020 IEEE/RSJ International Conference on Intelligent Robots and
  Systems (IROS), IEEE (2020)  10758--10765

\bibitem{hornung2013octomap}
Hornung, A., Wurm, K.M., Bennewitz, M., Stachniss, C., Burgard, W.:
\newblock Octomap: An efficient probabilistic 3d mapping framework based on
  octrees.
\newblock Autonomous robots \textbf{34}(3) (2013)  189--206

\bibitem{li2021multi}
Li, S., Chen, X., Liu, Y., Dai, D., Stachniss, C., Gall, J.:
\newblock Multi-scale interaction for real-time lidar data segmentation on an
  embedded platform.
\newblock IEEE Robotics and Automation Letters \textbf{7}(2) (2021)  738--745

\bibitem{wong2020identifying}
Wong, K., Wang, S., Ren, M., Liang, M., Urtasun, R.:
\newblock Identifying unknown instances for autonomous driving.
\newblock In: Conference on Robot Learning, PMLR (2020)  384--393

\bibitem{blum2022self}
Blum, H., Milano, F., Zurbr{\"u}gg, R., Siegwart, R., Cadena, C., Gawel, A.:
\newblock Self-improving semantic perception for indoor localisation.
\newblock In: Conference on Robot Learning, PMLR (2022)  1211--1222

\bibitem{wang2021unsupervised}
Wang, G., Tian, X., Ding, R., Wang, H.:
\newblock Unsupervised learning of 3d scene flow from monocular camera.
\newblock In: 2021 IEEE International Conference on Robotics and Automation
  (ICRA), IEEE (2021)  4325--4331

\bibitem{dewan2016rigid}
Dewan, A., Caselitz, T., Tipaldi, G.D., Burgard, W.:
\newblock Rigid scene flow for 3d lidar scans.
\newblock In: 2016 IEEE/RSJ International Conference on Intelligent Robots and
  Systems (IROS), IEEE (2016)  1765--1770

\bibitem{schaefer2019long}
Schaefer, A., B{\"u}scher, D., Vertens, J., Luft, L., Burgard, W.:
\newblock Long-term urban vehicle localization using pole landmarks extracted
  from 3-d lidar scans.
\newblock In: 2019 European Conference on Mobile Robots (ECMR), IEEE (2019)
  1--7

\bibitem{rusu2007towards}
Rusu, R.B., Blodow, N., Marton, Z., Soos, A., Beetz, M.:
\newblock Towards 3d object maps for autonomous household robots.
\newblock In: 2007 IEEE/RSJ International Conference on Intelligent Robots and
  Systems, IEEE (2007)  3191--3198

\bibitem{steininger2021density}
Steininger, M., Kobs, K., Davidson, P., Krause, A., Hotho, A.:
\newblock Density-based weighting for imbalanced regression.
\newblock Machine Learning \textbf{110}(8) (2021)  2187--2211

\bibitem{krawczyk2016learning}
Krawczyk, B.:
\newblock Learning from imbalanced data: open challenges and future directions.
\newblock Progress in Artificial Intelligence \textbf{5}(4) (2016)  221--232

\bibitem{carlevaris2016university}
Carlevaris-Bianco, N., Ushani, A.K., Eustice, R.M.:
\newblock University of michigan north campus long-term vision and lidar
  dataset.
\newblock The International Journal of Robotics Research \textbf{35}(9) (2016)
  1023--1035

\bibitem{xu2022fast}
Xu, W., Cai, Y., He, D., Lin, J., Zhang, F.:
\newblock Fast-lio2: Fast direct lidar-inertial odometry.
\newblock IEEE Transactions on Robotics (2022)

\bibitem{zou2013optimal}
Zou, K.H., Yu, C.R., Liu, K., Carlsson, M.O., Cabrera, J.:
\newblock Optimal thresholds by maximizing or minimizing various metrics via
  roc-type analysis.
\newblock Academic radiology \textbf{20}(7) (2013)  807--815

\bibitem{lawson2001geometric}
Lawson, J.D., Lim, Y.:
\newblock The geometric mean, matrices, metrics, and more.
\newblock The American Mathematical Monthly \textbf{108}(9) (2001)  797--812

\end{thebibliography}

\end{document}